\documentclass[manuscript]{acmart}

\usepackage{graphicx}

\usepackage{multirow}

\usepackage{booktabs} 

\usepackage{longtable} 
\makeatletter
\def\nobreakhline{%
  \noalign{\ifnum0=`}\fi
    \penalty\@M
    \futurelet\@let@token\LT@@nobreakhline}
\def\LT@@nobreakhline{%
  \ifx\@let@token\hline
    \global\let\@gtempa\@gobble
    \gdef\LT@sep{\penalty\@M\vskip\doublerulesep}
  \else
    \global\let\@gtempa\@empty
    \gdef\LT@sep{\penalty\@M\vskip-\arrayrulewidth}
  \fi
  \ifnum0=`{\fi}%
  \multispan\LT@cols
     \unskip\leaders\hrule\@height\arrayrulewidth\hfill\cr
  \noalign{\LT@sep}%
  \multispan\LT@cols
     \unskip\leaders\hrule\@height\arrayrulewidth\hfill\cr
  \noalign{\penalty\@M}%
  \@gtempa}
\makeatother

\begin{document}

\title{Deep learning models for predictive maintenance: a survey, comparison, challenges and prospect}

\author{Oscar Serradilla}
\orcid{0000-0003-2557-9859}
\affiliation{%
  \institution{Mondragon Unibertsitatea}
  \department{Electronics and Computer Science}
  \streetaddress{Loramendi 4}
  \city{Mondragon}
  \state{Spain}
  \postcode{20500}
}
\email{oserradilla@mondragon.edu}

\author{Ekhi Zugasti}
\orcid{0000-0001-8506-5695}
\affiliation{%
  \institution{Mondragon Unibertsitatea}
  \department{Electronics and Computer Science}
  \streetaddress{Loramendi 4}
  \city{Mondragon}
  \state{Spain}
  \postcode{20500}
}
\email{ezugasti@mondragon.edu}

\author{Urko Zurutuza}
\orcid{0000-0003-3720-6048}
\affiliation{%
  \institution{Mondragon Unibertsitatea}
  \department{Electronics and Computer Science}
  \streetaddress{Loramendi 4}
  \city{Mondragon}
  \state{Spain}
  \postcode{20500}
}
\email{uzurutuza@mondragon.edu}

\begin{abstract}

Given the growing amount of industrial data spaces worldwide, deep learning solutions have become popular for predictive maintenance, which monitor assets to optimise maintenance tasks. Choosing the most suitable architecture for each use-case is complex given the number of examples found in literature. This work aims at facilitating this task by reviewing state-of-the-art deep learning architectures, and how they integrate with predictive maintenance stages to meet industrial companies’ requirements (i.e. anomaly detection, root cause analysis, remaining useful life estimation). They are categorised and compared in industrial applications, explaining how to fill their gaps. Finally, open challenges and future research paths are presented.

\end{abstract}



\maketitle


\section{Introduction}

In recent years, industry has risen attention on artificial intelligence and machine learning techniques due to their capacity of creating automatic models that handle the big amount of data currently collected, which is growing exponentially. The research trend of machine learning has switched to more complex models such as ensemble methods and deep learning given their higher accuracy dealing with bigger datasets. These methods have evolved due to the increase of computing power and the latter mainly due to the evolution of GPU-s, being deep learning one of the most researched topics nowadays. These models achieve state-of-the-art results in many fields such as intrusion detection system, computer vision or language processing.

Maintenance is defined by the norm EN 13306 \cite{din13306} as \textit{the combination of all technical, administrative and managerial actions during the life cycle of an item intended to retain it in, or restore it to, a state in which it can perform the required function}. Moreover, it defines three types of maintenance: improvement maintenance improves machine reliability, maintainability and safety while keeping the original function; preventive maintenance is performed before failures occur either in periodical or predictive ways and corrective maintenance replaces the defective/broken parts when machine stops working. Currently, most industrial companies rely on periodical and corrective maintenance strategies.

Nowadays, we are transitioning towards the fourth revolution denominated as Industry 4.0 (I4.0), which is based on cyber physical systems and industrial internet of things. It combines software, sensors and intelligent control units to improve industrial processes and fulfill their requirements \cite{Lukavc2015}. These techniques enable automatised predictive maintenance functions analysing massive amount of process and related data based on condition monitoring (CM). Predictive maintenance (PdM) is the most cost-optimal maintenance type given its potential to achieve an overall equipment effectiveness (OEE) \cite{Vorne2019} higher than 90\% by anticipating maintenance requirements \cite{colemen2017,sanger2017} and promise a return on investment up to 1000\% \cite{Lavi2018Jul}. Maintenance optimisation is a priority for industrial companies given that effective maintenance can reduce their cost up to 60\% by correcting failures of machines, systems and people \cite{dhillon2002engineering}. Concretely, PdM maximises components' working life by taking advantage of their unexploited lifetime potential while reducing downtime and replacement costs by replacement before failures occur; thus preventing expensive breakdowns and production time loss caused by unexpected stops.

The numerous research works on PdM can be classified in three approaches \cite{liao2016hybrid}: physical model-based, data-driven and hybrid. Physical model methods use systems' knowledge to build a mathematical description of their degradation \cite{li2000stochastic, oppenheimer2002physically, venkatasubramanian2003review, blancke2018predictive, li2017expert}. It is easy to understand their physical meaning but difficult to implement in complex systems. Data-driven methods predict systems' state by monitoring their condition with solutions that learned from historical data \cite{baptista2018forecasting, zhang2005integrated, yuan2016fault}. These are composed of statistical methods, reliability functions and artificial intelligence methods. They are suitable for complex systems since they do not need to understand how these work. However, it is more difficult to relate their output to physical meaning. Finally, hybrid approach combines the aforementioned two approaches \cite{liao2016hybrid, zhao2013uncertainty}. Data-driven and deep learning methods have gained popularity in industry in recent years due to the increase of machine data collection, which enables the development of accurate PdM models in complex systems.

The \textit{review methodology} of this survey on \textbf{deep learning models application for predictive maintenance} is explained in this paragraph. First, context and applications of PdM are analysed. After that, different types of models are researched. Then, data-driven models are analysed. Finally, deep learning models are thoroughly reviewed. This methodology has enabled to acquire general insight of the scope and then focus on the specific research topics. Furthermore, the state-of-the-art (SotA) analysis has enabled the comparison among methods and discussion on challenges and prospect of DL models for PdM. The conducted analysis is performed by querying search engines about aforementioned topics. Initially, Scopus and Engineering Village search engines were used, since these contain more specific and relevant articles of the field. However, when the research advanced to more specific topics, another search engine was included: Google Scholar. This extends the research space to unindexed journals and preprints, providing a wider space including newer and unindexed published works. Many works belong but are not limited to the following publishing editorials: ACM Digital Library, ScienceDirect, IEEE-Xplore and SpringerLink.

Despite existing several published reviews on machine learning and deep learning models for predictive maintenance, this work provides these additional \textit{contributions} to the state-of-the-art (SotA): (1) We review and explain the most relevant data-driven techniques focused on SotA deep learning architectures with application to PdM, providing extensive perspective on the available techniques in a simplified and structured way. (2) We discuss the suitability of deep learning models for PdM and compare their benefits and drawbacks with statistical and classical machine learning models. (3) We analyse current trends on PdM publications, define their gaps, present research challenges, identifying opportunities and prospect.

This paragraph describes the remaining content of this work. Section 2 reviews predictive maintenance's background stages and provides an overview of traditional data-driven models used in the field, together with an overview about deep learning techniques. Section 3 reviews and categorises the most relevant state-of-the-art deep learning works for predictive maintenance by underlying technique, analysing them by PdM stages to enable comparison. Moreover, related reviews are analysed. Section 4 reviews the publicly available reference datasets for PdM model application and benchmarking. Section 5 discusses the suitability of deep learning models for predictive maintenance, evaluating their benefits and drawbacks in comparison with other data-driven techniques. Finally, Section 6 concludes this work by highlighting the most relevant aspects and gaps discovered during the review of referenced publications.

\section{Overview of predictive maintenance and deep learning}

\subsection{Predictive maintenance background}

Predictive maintenance solutions have to deal with many factors, peculiarities and challenges of industrial data. The most relevant ones are discussed in the next paragraphs.

Venkatasubramanian et al. present in \cite{venkatasubramanian2003review} the 10 desirable properties for a PdM system: \textit{quick detection and diagnosis, isolability (distinguish among different failure types), robustness, novelty identifiability, classification error estimation, adaptability, explanation facility, minimal modelling requirements, real-time computation and storage handling, multiple fault identifiability}.

Two main challenges of industrial use-cases are their behaviour and data variability. These occur even in assets working under same characteristics given the mechanical tolerances, mount adjustments, variations in EOC and other factors. These factors make PdM model reusability difficult among machines and assets. Other relevant challenges are gathering quality data, performing correct preprocessing and feature engineering to get a representative dataset for the problem. In addition, each observation is related to previous ones and therefore should be analysed together, which increases data dimensionality and modelling complexity; and failure data gathering is difficult given machines are designed and controlled to work correctly while preventing failures, therefore these are not frequent.

Some commonly monitored key components in PdM are but not limited to, bearings, blades, engines, valves, gears and cutting tools \cite{Zhang2019}. Moreover, the most common failure types detected by CM are imbalance cracks, fatigue, abrasive and corrosion wear, rubbing, defects and leak detection among others. The publication by Li et al. \cite{Li2010} classifies the types of failures that may exist in the system as: component failure, environmental impact, human mistakes and procedure handling.

The commonly used CM techniques are the following ones \cite{UESystems2019}: mechanical ultrasound \cite{bakar2013PdMUltrasoundThermal}, vibration analysis \cite{amruthnath2018research, Durbhaka2016, wen2018degradationKDE, feng2012PermutationEntropy, saxena2013assessment}, wear particle testing \cite{Woldman2015PdMWearParticles, ELNASHARTY2011588}, thermography, motor signal current analysis \cite{Santos2017MSCA} and nondestructive testing \cite{drinkwater2009ultrasonic}, but there are additional techniques as torque, voltage and envelopes \cite{mehrjou2011rotor}, acoustic emission \cite{saxena2013assessment},  pressure \cite{zhao2017PdMAD_CorrelationADtemperaturePressure} or temperature monitoring \cite{bakar2013PdMUltrasoundThermal, zhao2017PdMAD_CorrelationADtemperaturePressure}. The articles \cite{selcuk2017predictive, marquez2012condition} also dive into these techniques and cover the types of failures they can detect, together with their applications. They highlight that EOC information could complement these CM techniques to perform a more robust PdM analysis, collecting data from different sources: physical, machine and operating.

Environmental and operational conditions (EOC) are conditions under which an industrial asset such a machine or component is working \cite{Tavner2008}. Environmental conditions refer to external conditions that affect them like ambient temperature or surrounding vibration perturbations. In contrast, operational conditions are working processes' assigned technical specifications, such as desired speed, force or positions. Additionally, sensor data comes from measurements taken by machine sensors. This data monitored over the time creates a dataset, in a form of time-series. Its analysis using condition monitoring techniques enable determining component and machine states by comparing patterns and trends with historical data. Many works present component degradation patterns in a plot denominated as P-F curve \cite{UESystems2019}, where health decreases from healthy working condition until failure as time or machine cycles go by.

\subsection{Data-driven predictive maintenance stages} \label{sec:data-driven-PdM-stages}

\begin{figure}[hb]
    \centering
    {\includegraphics[width=8cm]{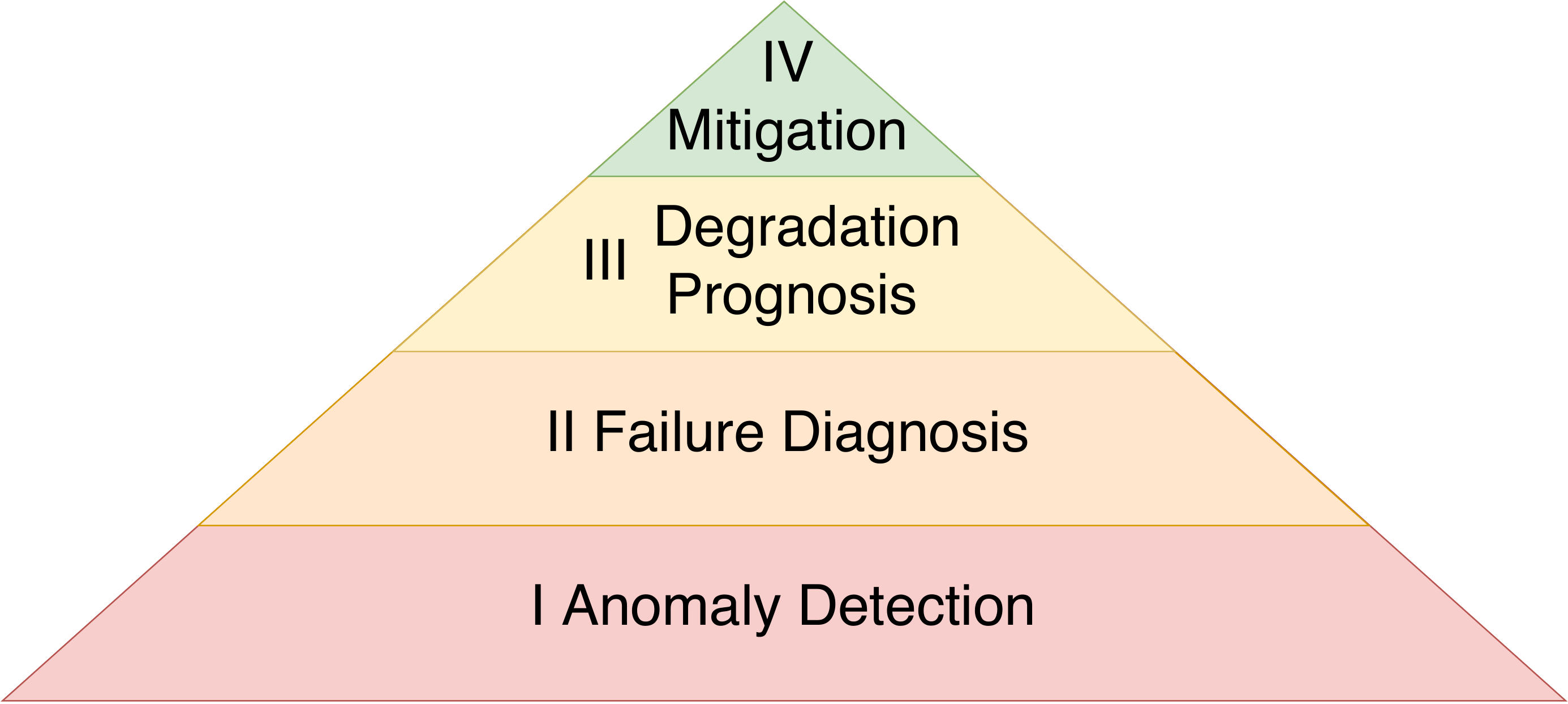}}
    \caption{Predictive maintenance roadmap represented by a stages of a pyramid chart.}
    \label{fig:PdM-roadmap-diagram}
\end{figure}

The majority of deep learning models for PdM are based on the same principles as other machine learning and statistical techniques. Precisely, most data-driven methods follow the incremental steps presented in the roadmap of Figure \ref{fig:PdM-roadmap-diagram}, based on the articles \cite{Welz2017, prajapati2012condition} and OSA-CBM standard \cite{lebold2002osa}: 1st \textbf{anomaly detection}, 2nd \textbf{diagnosis}, 3rd \textbf{prognosis} and lastly \textbf{mitigation}.

Commonly two additional steps are performed before the aforementioned ones to prepare the data for PdM, as general data analytic lifecycle, Khan et al. \cite{khan2018review} and other PdM authors present. These steps are preprocessing and Feature Engineering (FE), which, as stated above, are key to enhance model accuracy on PdM stages by creating a representative dataset for the problem. All PdM stages have to be designed, adapted and implemented to fit use-cases' requirements and their data characteristics. In addition, the PdM systems development is incremental and therefore, techniques, algorithms and decisions taken in each stage will influence the following ones. The next subsections overview the most common data-driven methods to address each PdM stage.

\subsubsection{Preprocessing} This step consists of preparing the collected data for further stages. Each PdM model has different requirements and these must be taken into consideration when choosing adequate preprocessing techniques to boost model performance. The most common techniques are briefly explained and referenced below: sensor data validation \cite{Zugasti2012} makes sure the collected data is correct; feature synchronisation \cite{Kroll2014} is used to gather signals sampled at different timestamps to create a time-series/cycle-based data that is easier to handle; data cleaning removes or interpolates not available and missing values \cite{cerqueira2016LOFPdM, costa2016IDAPdMChallenge}; oversampling \cite{cerqueira2016LOFPdM, perini2019PdMthesisHMM_kNN_IF_AE} is applied for imbalance data handling to boost accuracy on commonly scarce failure data class or to deal with small datasets; encoding \cite{Martinez} and discretisation \cite{Martinez} change features' type by projection to a new space where they are easier to handle by the model; segmentation splits data in chunks to analyse big datasets and enable parallelisation \cite{Lo2018}; feature scaling like normalisation \cite{Sanayha2017AD_ARIMAandMahalanobis} or standardisation \cite{pratama2018online} scales all features to the same or similar space that enables comparisons; noise handling \cite{Kroll2014} facilitates noisy data modelling.

\subsubsection{Feature engineering} This step consists of extracting a relevant feature subset to be used as input for models in further stages. It can boost statistical and machine learning model performance, despite not being compulsory for deep learning models given these can extract new representative features that fit the problem automatically. The most common techniques can be grouped into next groups: feature extraction as statistical features in time \cite{Zhang2019} and frequency \cite{Zhang2019, zhou2014SVMPdM, GARG2015GPPdMFFT} domains that extract time/frequency relations of features; based on projection to new space like principal component analysis \cite{Santos2017MSCA, canizo2017PdMWindTurbine} which reduce dimensionality while keeping relevant information; concatenation and fusion methods \cite{Lee2006} create new features by combining available ones; feature selection \cite{Serradilla2020} reduces dimensionality discarding features of low variance, redundant and  uncorrelated to target, given these increase complexity while not supplying additional information.

\subsubsection{Anomaly detection} It aims to detect whether the asset is working under normal condition or not. There are three ways to address this step using data-driven models, classified by their underlying machine learning task: classification, one-class classification and clustering. Respectively, these can be used when labeled data of different classes is available in the training phase, when only one class data exist (commonly non-failure data) and when the data is unlabelled. Failure modes and effects analysis (FMEA) \cite{dhillon1992failure} and its evolution by adding criticality analysis FMECA \cite{jordan1972failure} are useful to gain vision on the possible types of failures based on expert knowledge, which helps designing the data analysis lifecycle, prioritising the failure types or anomalies to be detected.

The anomaly detection methods need preprocessed and some also depend on feature engineered data to work. Once worked on features, the next step is to select, train and optimise the right model for the use-case. Following PdM stages will be influenced and constrained by the selected AD method and use-case's data. Table \ref{tab:AD_mod_sum} classifies and summarises the main data-driven anomaly detection techniques based on referenced SotA articles and the following review works \cite{Wang2019, Zhang2019, pimentel2014review, carvalho2019systematic}. Besides, two or more of these techniques can be combined to create an anomaly detection system that compensates the disadvantages of a single model.

\begin{table}[ht]
\caption{Summary of anomaly detection models classified by prevailing techniques. In the first column, Unsup refers to unsupervised, All refers to supervised, semi-supervised and unsupervised and and Combination refers to a combination of models respectively.}
\label{tab:AD_mod_sum}
\hfill
\small 
\begin{minipage}{\columnwidth}
\begin{tabular}{p{10mm}p{15mm}p{10mm}p{15mm}p{83mm}}

\hline
\textbf{Based on and Type} & \textbf{What analyses} & \textbf{Normal data} & \textbf{Anomalies} & \textbf{Most common algorithms and categorised} \\ \hline

Density Unsup & Density in features space dimension & In high density & In low density & K nearest neighbors (k-NN) \cite{susto2014machine, mathew2017prediction, costa2016IDAPdMChallenge, perini2019PdMthesisHMM_kNN_IF_AE}, local outlier factor (LOF) \cite{cerqueira2016LOFPdM, DiezOlivan2017}, local correlation integral (LOCI) \footnote{Methods that have been applied for AD in general but not specifically for PdM are mentioned but not referenced}, relative density factor, density-based outlier score, reliability functions \cite{Zhai2019reliabilityFunction, okoh2017PdMReliabilityFunction} \\ 
Distance Unsup & Distance among data-points & Near from neighbors & Far from neighbors & Traditional threshold distance mahalanobis \cite{Sanayha2017AD_ARIMAandMahalanobis} or euclidean \cite{Durbhaka2016}, rank based detection algorithm (RBDA), randomization and pruning based, data streams based \\ 
Statistics All & Relation to distribution models fit to training data & Near to distribution models & Far from distribution models & \textbf{Parametric}: gaussian mixture models (GMM) with expectation maximisation (EM) \cite{amruthnath2018research}, control charts as exponentially weighted moving average (EWMA) \cite{castagliola2011EWMAforPdM, susto2012PdM_EWMA_KDE}. \textbf{Non-Parametric}: kernel density estimation (KDE): gaussian or KL-divergence \cite{Zhao2010PdM_KDE, susto2012PdM_EWMA_KDE, wen2018degradationKDE}, histogram-based outlier detection (HBOS) \cite{Munir2017HBOSforPdM}, boxplot analysis \cite{cerqueira2016LOFPdM}, 3$\sigma$ \cite{able2016PdM_3sigma}.
\textbf{Entropy-based} permutation entropy \cite{feng2012PermutationEntropy, Radhakrishnan2016PermutationEntropy}, fuzzy entropy \cite{Carlsson2016PdMFuzzyEntropy} and K-S test \cite{bolbolamiri2012time}. \\ 
Clustering Unsup & Relation to clusters created by unsupervised ML models & Belong to a large cluster or near one & Belong to a small cluster and far from large clusters & \textbf{Partitioning clustering}: partitioning around medoids (PAM), K-means \cite{DiezOlivan2017, Eke2017298, amruthnath2018research}. \textbf{Hierarchical clustering}: DB-Scan, agglomerative \cite{cerqueira2016LOFPdM}, attribute oriented induction (AOI) \cite{fernandez2019attribute}. \textbf{Grid-based}: Dcluster. \textbf{For high dimensional}: D-Stream, fuzzy-rules based \cite{DiezOlivan2017} \\ 
Ensemble Combination & Combines dissimilar models. Robust & Combina-tion of models & Combination of models & Bagging or boosting based as random forest (RF) \cite{mathew2017prediction, costa2016IDAPdMChallenge, canizo2017PdMWindTurbine, Santos2017MSCA}, extra gradient boosting (XGBoost) \cite{cerqueira2016LOFPdM}, adaboost \cite{mathew2017prediction}  and isolation forest (IF) \cite{perini2019PdMthesisHMM_kNN_IF_AE}, greedy ensemble, score normalization \\ 
Learning All & Relation to models learned with training data & Near the known classes of the model & Far from the known classes of the model & \textbf{Active learning}. \textbf{Transfer learning}. \textbf{Reinforcement learning}. \textbf{Projection-based}: Subspace and compression reconstruction error measuring like PCA \cite{amruthnath2018research} and AE \cite{chen2017deep}, correlation \cite{zhao2017PdMAD_CorrelationADtemperaturePressure, zugasti2018null} and tensor-based. \textbf{State-space based} (hidden state of observed data and time evolution): kalman filter \cite{Vianna2018}, hidden markov models (HMM) \cite{perini2019PdMthesisHMM_kNN_IF_AE}, bayesian networks (BN) \cite{kolokas2018NaiveBayesPdM} (dynamic BN, belief network), attention-based NN and RNN (GRU, LSTM). \textbf{Graph-based}: capture interdependiencies. \textbf{OCC}: OCSVM \cite{zhou2014SVMPdM}, BN. \textbf{Prediction error-based regression}: measure deviation (autoregressive integrated moving average (ARIMA) \cite{Sanayha2017AD_ARIMAandMahalanobis}, RNN as LSTM \cite{yuan2018lstm}). \textbf{Classification}: normal and abnormal data in training using \underline{interpretable models}: linear regression \cite{mathew2017prediction}, logistic regression \cite{mathew2017prediction, costa2016IDAPdMChallenge}, decision tree (DT) \cite{costa2016IDAPdMChallenge, Jegadeeshwaran2015SVMPdM}. \underline{ML classification} techniques as SVM \cite{costa2016IDAPdMChallenge, Jegadeeshwaran2015SVMPdM, susto2014machine} and feedforward NN \cite{rad2011artificial}. \textbf{Generative methods}: GAN \cite{lee2017application}, VAE \cite{xu2018unsupervised}.
\\ \hline

\end{tabular}

\end{minipage}

\end{table}

\subsubsection{Diagnosis}

Once an anomaly has been detected, the next stage consists of diagnosing whether this anomaly belongs to a faulty working condition and can evolve into a future failure or, in contrary, there is no risk of failure. The last case indicates that the anomaly detection model has not worked properly and therefore it may need to be reevaluated or retrained. The diagnosis is usually based on root cause analysis (RCA) techniques, which aim to identify the true cause of a problem. 

The diagnosis algorithm has to be suitable for the problem being addressed. There are several approaches to tackle this step, which depend on the implemented AD method and training data characteristics: multi-class classification, binary classification, one-class classification and clustering. Concretely these are chosen if the dataset has multiple failure types, failure and non failure observations, only observations of one class or unsupervised, respectively. There is another technique that commonly complements RCA: anomaly deviation quantification by health index (HI). It aims to measure assets' damage by comparing current working data with historical data in a supervised or unsupervised way. It can either indicate a percentage of deviation with regard to normal working data, or show degradation level in a numerical scale, where the higher the value the more damaged the component is, where minimum value means no damage, maximum is fully damaged or failure and intermediate values indicate different degrees of degradation \cite{lifetimeRel2019OctHI}.

The diagnosis step is easier when there is more information about the dataset and its labels. The main statistical and machine learning techniques for diagnosis are described in the following list, ordered by increasing difficulty. They are divided according to the anomaly detection technique used in the previous stage, which depends on data characteristics.

\begin{itemize}
    \item After multi-class classification for anomaly detection: diagnosis is performed based on previous failure data knowledge of the estimated class, so the link of data to failure type is directly obtained from model \cite{bakar2013PdMUltrasoundThermal, boutros2011HMMforDiagnosis}. Once the possible failure type has been detected, semi-quantitative and qualitative approaches can be used by harnessing expert knowledge to evaluate its potential consequences, using tools such as FMEA \cite{cortes2016strategic} or Ishikawa diagram \cite{pradhan2007bayesian}. In addition, interpreting directly explainable models \cite{ansari2020DynamicBNforPdM, Ademujimi2017RCAsummary} or using explainability on less interpretable models such as SVM \cite{demetgul2013fault} can also help to perform this task.
    \item After binary classification for anomaly detection: clustering with extracted features can be performed to group data by similarity and try to differentiate unlabeled failure types \cite{yuwono2016automatic}. These diagnosis techniques can also be based on statistical performance analysis \cite{Murugan2015RCAstatistics}, supported on trend analysis and definition of thresholds to differentiate failure types by similarity or distance.
    \item After one-class classification or clustering for anomaly detection: these techniques use a threshold in distance to the classified class or clusters density respectively to categorise anomalies. Diagnosis for these models usually consists of precomputing metrics from data like health index and monitoring their evolution, instead of monitoring input data evolution. The diagnosis can be performed using a clustering algorithm in these metrics to analyse the intra-cluster and inter-cluster relations. Domain knowledge is essential to tie unsupervisedly discovered relations to physical meaning of monitored assets. This novel knowledge is useful for interpreting unsupervised models' output to discover novel failure types, using models as K-means \cite{adhikari2018machine} or HMM with IF-ELSE rules \cite{wu2018clusterHMMforRCAandRUL}. Log data can also be used for this clustering purpose and tag maintenance data \cite{sharp2017toward} to perform RCA.
\end{itemize}

\subsubsection{Prognosis} Once an anomaly is detected and diagnosed, the degradation evolution can be monitored based on that moment's working conditions and machine state, focusing on the most influential features for AD and diagnosis stages that can track failures. This step is usually carried out by remaining useful life models that estimate the remaining time or cycles until a failure occurs when there is enough historical data of that failure type. Conversely, if there is not enough degradation data, the only way to estimate degradation is by tracking the evolution of HI or the distance between novel working states and the known good working states. Both aforementioned models can also provide a confidence bound. The data-driven models for prognosis can be classified into 4 groups regarding their underlying method. The following list summarises the most common techniques categorised by groups to prognosticate degradation:

\begin{itemize}
    \item Similarity-based: compare current behavior with past run-to-failure behavior for prognosis \cite{adhikari2018machine, ramasso2014investigating}.
    \item Statistical: rely on historical statistics to estimate degradation, for example monitoring life usage in combination with mean-time-to-failure \cite{baptista2018forecasting} or survival models \cite{zhang2015seeing} to estimate the expected duration.
    \item Time series analysis: ARIMA \cite{Kanawaday2017ArimaPdM, adhikari2018machine, baptista2018forecasting} based on previous values, kalman filter to model hidden state of time-related noisy data \cite{Vianna2018} and fourier and genetic programming to generate a polynomial function by optimising a fitness function \cite{GARG2015GPPdMFFT}.
    \item Learning:
    \begin{itemize}
        \item Classification: diagnose the data to a known failure type or similar working data and then prognosticate a degradation according to the historical data of this class. Despite any classifier can be used for this purpose, the following ones are widely used in literature: feed-forward NN \cite{rad2011artificial}, SVM \cite{rad2011artificial}, BN \cite{lee2019evaluating, lee2017predictive, ansari2020DynamicBNforPdM}, HMM \cite{zhang2005integrated}, fuzzy logic based \cite{ZIO201049} and RF \cite{Gutschi2019, baptista2018forecasting}.
        \item Regression: directly estimate HI, anomaly deviation or RUL from the input data. Common SotA algorithms are below: linear function is the simplest method \cite{sharp2017toward}; nonlinear functions \cite{liao2006logisticRegressionRUL, zhang2015seeing} can model non-linear relations; support vector regressor (SVR) \cite{Benkedjouh2013RUL_SVR, baptista2018forecasting} works like SVM adapted for regression; relevance vector regression (RVR) is based on bayesian regression \cite{adhikari2018machine}; CNN models features' time-based relationships \cite{babu2016RULwithCNN}; wiener processes model degradation by a real valued continuous-time stochastic processes \cite{si2013wiener}; recurrent neural networks like LSTM and GRU \cite{yuan2016fault} retain relevant past information for prognosis at each observation.
    \end{itemize}
\end{itemize}

\subsubsection{Mitigation} Once an anomaly is detected, diagnosed its cause and prognosticated its remaining life, there is enough information to perform maintenance actions to mitigate failures in early phases and thus prevent assets deriving into failure. This stage consists of designing and performing the steps necessary to restore assets to correct working condition before failures occur, which also reduces implementation and downtime costs.

Mitigation is performed by maintenance technicians who are in charge of creating and implementing a mitigation plan as part of the maintenance management and manufacturing operation management processes. Data-driven PdM models should generate assistance information, providing domain technicians with statistics \cite{Murugan2015RCAstatistics} and prescriptions \cite{ansari2020DynamicBNforPdM}. Therefore, a more advanced mitigation is accomplished by the combination of domain knowledge and data-driven information about assets' health and expected degradation \cite{liu2012condition}.

\subsection{Deep learning techniques}

This section presents deep learning background and introduces its most common architectures applied to the field of PdM. Nowadays, deep learning models outperform statistical and traditional ML models in many fields including PdM, when enough historical data exist. The deep learning (DL) term refers to artificial neural networks (ANN), a machine learning technique inspired on brain functioning, that \textit{go beyond shallow 1- and 2-hidden layer networks} \cite{nielsen2015neural}.

ANNs are formed by neurons that compute linear regressions of inputs with weights and then compute non-linear activation functions such as sigmoid, rectified linear unit (reLU) or tan-h to produce outputs. The network's parameters are commonly initialised randomly and they are then adjusted to map input data to output data given the training dataset. This learning process takes place by running gradient descend algorithm combined with backpropagation algorithm. These enable to calculate the adjustments of each neuron with respect to the error produced by the network to reduce it, where the error is calculated based on the user defined cost function. Hornik in the article \cite{hornik1991approximation} justifies that ANNs of at least two hidden layers with enough training data are capable of modelling any function or behaviour, creating the universal approximator.

The book by Goodfellow et al. \cite{GoodfellowDeepLearningBook} provides exhaustive background on DL and it is considered as reference book by many researchers in the field. Concretely, the book introduces machine learning and deep learning mathematical background. Afterwards, it focuses on DL optimisation, regularisation, different type of architectures, their mathematical definition and common applications. A simpler yet powerful overview of the field is done by Litjens et al. in the survey of DL applied to medicine \cite{litjens2017survey}, which is further complemented with a visual scheme collecting the main architectures. Another survey specifically focused on DL architectures, applications, frameworks, SotA and historical works, trends and challenges is the one by Pouyanfar et al. \cite{pouyanfar2019survey}. Additionally, a reference book of practical DL applications is presented by Geron \cite{geron2019hands}, which is based on the following tools: Scikit-Learn~\footnote{https://scikit-learn.org}, Keras~\footnote{https://keras.io}, and TensorFlow~\footnote{https://www.tensorflow.org}.

The most common DL \textit{techniques} related to the field of PdM are summarised in the following paragraphs. Most of them are based on the feed forward scheme but each one has its own characteristics:

\begin{itemize}
    \item Feed-forward/MLP \cite{werbos1982applications} is the first, most common and simplest architecture. It is formed by stacked neurons creating layers, where all the neurons of a layer are connected to all the neurons of the next layer by feeding their output to others' input. However, there are no connections to neurons of previous layers or among neurons of the same layer. The nomenclature for layers is the following: an input layer, hidden layers and an output layer. The neural network is fed with observations pairing input features and target features, which are used to learn their relation by minimising the error produced by the network by mapping input data to output.
    \item Convolutional neural network (CNN) \cite{lecun1989backpropagation} is a type of feedforward network that maintains neurons' neighborhood by applying convolutional filters. It is inspired by the animal visual cortex and has applications in image and signal recognition, recommendation systems and NLP among others. The convolutional layer is usually linear and is followed by the application of an activation function to produce non-linear output. After that, a max or average pooling layer can be used to reduce the dimension. Finally, most architectures have a flatten step to obtain representative features of input data that can be used with other ML or DL networks to perform typical ML tasks. The convolutions' weights are shared, making them easier to train.
    \item Recurrent neural network (RNN) \cite{robinson1987utility} models temporal data by saving the state derived from previous inputs of the network. The back-propagation through time algorithm \cite{werbos1988generalization} is an adaptation of traditional backpropagation for temporal data used to propagate network's error to previous time instances. However, this propagation can result into vanishing or exploding gradient problem \cite{hochreiter1991untersuchungen}, making this networks forget long-term relations. To solve this problem, specific RNN architectures were created based on forget gates, like long-short term memory (LSTM) \cite{hochreiter1997long} and gated recurrent unit (GRU) \cite{cho2014properties}.
    \item Deep belief network (DBN) \cite{hinton2006reducing} and restricted boltzmann machine (RBM) \cite{smolensky1986parallel}. RBM is a bipartite, fully-connected, undirected graph consisting of a visible layer and a hidden layer. It is a type of stochastic ANN that can learn probability distribution over the data. It can be trained in supervised or unsupervised ways and its main applications are on dimensionality reduction and classification. Accordingly, DBN is an ANN where every two consecutive layers are treated as RBMs. It is trained in unsupervised way to reduce dimensionality. Then, it can be retrained with classified data to perform classification.
    \item Autoencoder (AE) \cite{ballard1987modular} is based on singular value decomposition concept \cite{golub1971singular} to extract the non-linear features that best represent the input data in a smaller space. It consists of two parts: an encoder that maps input data to the encoded, latent space, and the decoder, which projects latent space data to the reconstructed space that has the same dimension as input data. The network is trained to minimise the reconstruction error, which is the loss between input and output. Autoencoders can be classified according to their latent space dimensionality in undercomplete and overcomplete, which respectively correspond to a latent space smaller, and bigger or equal to the input dimension. These simple architectures are extended and adapted to fit different tasks and problems. Vanilla autoencoders are the simplest autoencoders, which belong to the undercomplete type. The following variations are obtained by applying regularisation and modifying AE types. One of these adaptations is the denoising autoencoder (DAE) \cite{le1987modeles}, used for corrupt data reconstruction. It is a type of overcomplete AE where learning is controlled to avoid \textit{"identity function"}. It is fed with data pairs of noisy input and its denoised output and trained to reduce the loss between them. Another modification is the sparse autoencoder (SAE) \cite{makhzani2013k}, an AE restricted in the learning phase based on a sparse penalty constraint, which is based on the concept of KL-Divergence. This algorithm aims to make each neuron sparse, discovering the structure information from the data easier than vanilla AE and being more useful for practical applications \cite{sun2017intelligent}.
    \item Generative models: variational autoenconder (VAE) \cite{diederik2014auto} and generative adversarial network (GAN) \cite{goodfellow2014generative}. Both models were designed to work in unsupervised way. VAE is a generative and therefore non-deterministic modification of the vanilla AE where the latent space is continuous. Usually, its latent space distribution is gaussian, from where the decoder reconstructs the original signal based on random sampling and interpolation. It has applications on estimating the data distribution, learning a representation of data samples and generating synthetic samples among others. GAN is another type of generative neural network that consists of two parts: generator and discriminator. The generator is trained to generate an output that belongs to a specific data distribution using as input a representation vector. The discriminator is trained to classify its input data whether it belongs to a specific data distribution or not. The generator's output is connected to discriminator's input and they are trained together, adversely. Generator's objective is to bias discriminator by generating outputs from random input and trying to make the discriminator classify it as it belongs to the specific trained distribution. The role of the discriminator is to distinguish between synthetic, generated data, from non-synthetic, real data from trained distribution. They are trained together so that each part learns from the other, competing to bias the other part, similar to game theory. GANs can be extended to other ML tasks such as supervised or reinforcement learning.
    \item Self organising map (SOM) \cite{Kohonen1990SOMarticle} is an ANN-based unsupervised way to organise the internal representations of data. It uses competitive learning, in contrast to typical ANNs that use backpropagation and gradient-descend, to create a new space called map that is typically 2 dimensional. It is based on neighborhood functions to preserve the topological properties of the input space into the new space, represented in cells. It has applications on clustering, among others.
\end{itemize}

\section{Deep learning for predictive maintenance}

This section collects, summarises, classifies and compares the reference DL techniques for PdM, analysing the most relevant works and applications. It contains accurate DL models that achieve SotA results on reviewed articles, surveys and reviews of the field. Even though most articles combine several techniques and perform more than one PdM stage in the same architecture, this section classifies in its five first subsections the works by their principal DL technique to perform each stage of Section \ref{sec:data-driven-PdM-stages}; including the previous stage feature engineering and excluding preprocessing given the latter's explanation on previous section is also valid for DL. This classification enables the analysis and comparison of DL techniques by stages. The sixth subsection presents works that successfully combine the aforementioned techniques to create more complete architectures that fulfil one or more PdM stages, to give examples of ways to combine techniques that can be infinite. Finally, last subsection gathers the most relevant information contained in similar works to this survey, discussing related reviews and surveys.

The SotA works can be classified regarding their underlying ML task and algorithms used to address it, which are directly related to the use-case and its data requirements. Binary classification is used when training data contains labelled failure and non-failure observations. Multi-class classification is used in the same case as binary classification, but there is more than one type of failure classified and therefore there are at least three classes: one represents non-failure and then one for each type of failure. One-class classification (OCC) is used when the training dataset only contains non-failure data, which usually consists of collecting machine data in early working states or when technicians assure the asset is working correctly. Finally, unsupervised techniques are used when training datasets' observations are unlabelled and therefore there is no knowledge of which observations belong to failure and non-failure classes. Unsupervised techniques can also be used as one class classifiers. Additionally, there are a few works on other machine learning and deep learning topics such as active learning, reinforcement learning and transfer learning.

\subsection{Feature engineering}

The deep learning algorithms used in PdM are capable of performing feature engineering automatically, obtaining a subset of derived features that fit specifically for the task. Therefore, these techniques remove the dependence on manual and feature engineering process. Table \ref{tab:DL_for_feature-engineering} shows common deep learning techniques used for feature engineering. These techniques are integrated with machine learning and deep learning models to create architectures that perform PdM stages.

\begin{table}[!htbp]
\small 

\caption{Deep learning techniques for automatic feature engineering and projection. They are based on input signals relations and temporal context.}
\label{tab:DL_for_feature-engineering}

\begin{tabular}{p{12mm}p{29mm}p{33mm}p{33mm}p{16mm}p{7mm}}

\hline

\textbf{Algo-rithm} & \textbf{How it works} & \textbf{Strengths} & \textbf{Limitations} & \textbf{Applica-tions} & \textbf{Ref} \\ \hline
Feed-forward & Add deep layers with less dimensions & Reduce dimension to lower feature space. Simplest NN architecture & Does not model the features by neighborhood neither temporal relations. & Engine health monitoring, vibration monitoring & \cite{yildirim2016engine, rad2011artificial, al2010rolling} \\ 
RBM & Automatic feature extraction. Models data probability by minimising Contrastive Divergence. One-way training, reconstructing input from output. & Keep spatial representation in new space. Not much training time. & Not keeping data variance in new space. Difficulty on modelling complex data since only one layer. & Bearing degradation, factory PLC sensors & \cite{liao2016enhanced, hwang2018svm} \\ 
DBN & Automatic feature extraction using stacked RBMs with greedy training. Can be used for HI construction. & Competitive SotA results. Can model time-dependencies using sliding windows. & Very slow and inefficient training. Not modelling long-term dependencies. & Bearings vibration, aviation engine, wind turbine & \cite{wang2017hydraulic, deutsch2017using, peng2019deep, yang2018unsupervised, shao2017deep} \\ 
SOM & Data mapping to a specified dimension & Non-linear mapping of complex data to a lower dimension. Maintains feature distribution in the new space. Can be combined with other techniques for RCA (i.e. 5-whys \cite{chemweno2016development}) & Difficult to link latent variables with physical meaning. More complex than other techniques. Fixed number of clusters & Turbofan, pneumatic actuator, thermal power plant, bearing degradation & \cite{lacaille2015turbofan, prabakaran2014self, chemweno2016development, liao2016enhanced} \\ 
AEs & Dimensionality reduction in latent space keeping maximum input data variance. Non-linear FE and HI calculation. & Automatic FE of raw sensor data achieve similar results to traditional features. Traditional features can also be input. No need of classification or failure data. Allows online CM. & Extracted features not specific for the task. Needs more resources: computational and training data. Loses temporal relations if input data are raw sensors data. Can lead to overfitting & Bearing vibration, satellite data, PHM2012 Predictor Challenge & \cite{sakurada2014anomaly, chen2017deep, ahmed2018intelligent, ren2018bearing, perini2019PdMthesisHMM_kNN_IF_AE} \\ 
CNN & Automatic feature extraction. Univariate or multivariate convolutions of input. Models sequential data. Used with sliding windows. Combined with pooling methods to reduce dimension & Simple yet effective. Faster than traditional ML models in production. Takes advantage of neighborhoods. Less training time and data by weight-sharing. Can outperform LSTM. Dropout can prevent overfitting & Slower training due to high number of weights. Data analysis in chuncks, not modelling long-term dependencies. & Bearing, electric motor, turbofan & \cite{cabezas2019generative, li2018remaining, guo2016hierarchical, liu2016dislocated, wang2017virtualization, babu2016RULwithCNN, munir2018deepant, li2019novel} \\ 
RNNs & Regression. Model time-series and sequential data by propagating state information through time. & Model temporal relationships of EOC data. Special architectures as LSTM and GRU can model medium-term dependencies & RNNs suffer vanishing gradient problem, even special architectures cannot model very long-term dependencies. Need more resources. & Aero engine, hydropower plant & \cite{yuan2016fault, bruneo2019use, aydin2017using, yuan2018lstm} \\ \hline
\end{tabular}

\end{table}

\subsection{Anomaly detection}

\begin{figure}[!htbp]
    \centering
    {\includegraphics[width=15cm]{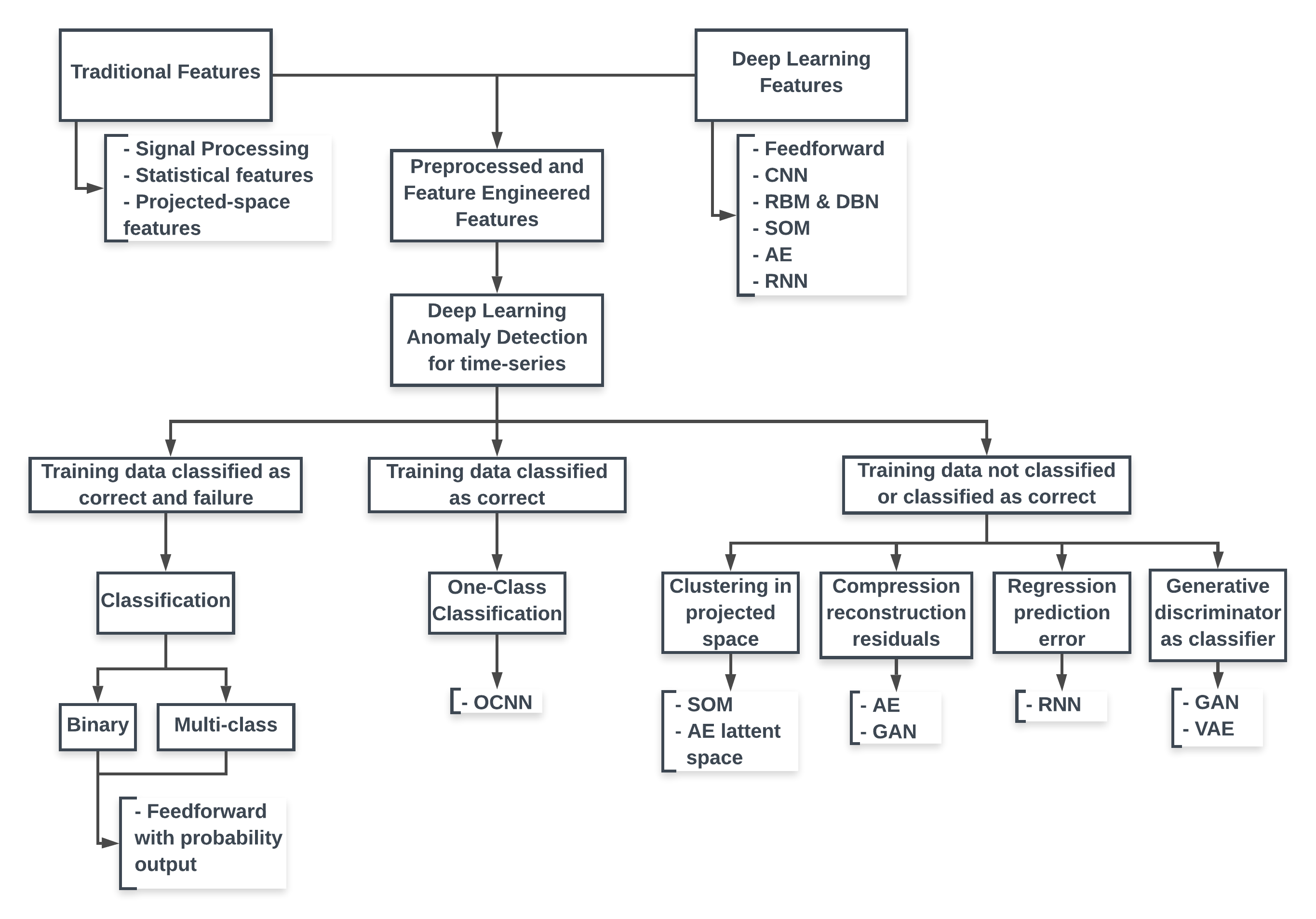}}
    \caption[caption for LOF]{Diagram of the main deep learning techniques for anomaly detection applied to predictive maintenance and time-series, classified by machine learning task\protect\footnotemark.}
    \label{fig:DL_AD_for_TS}
\end{figure}

The deep learning-based AD algorithms can be classified in three groups, as stated in the introduction of this section, regarding the characteristics of training data. The main architectures have been summarised in Figure \ref{fig:DL_AD_for_TS}.

\footnotetext{In this work, the term traditional features refers to handcrafted and automatic feature extraction techniques such as statistical or ML-based features, excluding DL-based features.}

Those algorithms are summarised, compared and their main applications are referenced in the following tables. On one hand, the anomaly detection algorithms based on binary and multi-class classification approaches \cite{rad2011artificial, al2010rolling} rely on training data classified as correct and failure. These commonly used feature extraction techniques either traditional or deep learning followed by a flatten process, and then have several fully-connected layers of decreasing dimension until the output layer. The output layer usually uses the softmax activation function to output the probability of failure and not failure. In the case of binary classification, there are one or two neurons indicating the probability of failure and normal working condition. Similarly, in multi-class classification there are N+1 number of neurons, where there is one neuron to indicate the probability of not failure and each of remaining N indicate the probability of each type of failure. On the other hand, Table \ref{tab:AD_oneclass_unsup} contains algorithms that address AD problem based on one-class classification or unsupervised approaches, using only training data classified as correct or not classified.

{\small 
\setlength\LTleft{0mm} 
\begin{longtable}{p{8mm}p{25mm}p{35mm}p{35mm}p{16mm}p{7mm}}
\caption{Anomaly detection methods that use training data classified as correct or not classified: one-class classification and unsupervised.} \label{tab:AD_oneclass_unsup} \\ \hline
\textbf{Algo-rithm} & \textbf{How it works} & \textbf{Strengths} & \textbf{Limitations} & \textbf{Applica-tions} & \textbf{Ref}
\endfirsthead
\nobreakhline
\textbf{Algo-rithm} & \textbf{How it works} & \textbf{Strengths} & \textbf{Limitations} & \textbf{Applica-tions} & \textbf{Ref}
\\* \nobreakhline
\endhead
\nobreakhline
\multicolumn{6}{l}{\textbf{Autoencoders}}\\*
Vani-lla AE & Threshold in reconstruction error. Data is considered when it surpasses the threshold.  & Automatic feature engineering of raw sensor data or traditional features. Minimise variance loss in latent space. No need of classification or failure data. Allows online CM & Extracted features not specific for the task. Needs more resources: computational and training data. Loses temporal relations if input data are raw sensors data. Can lead to overfitting & Bearing vibration, satellite data, PHM2012 Predictor Challenge & \cite{sakurada2014anomaly, chen2017deep, ren2018bearing, perini2019PdMthesisHMM_kNN_IF_AE, reddy2016anomaly} \\ 
Stack-ed AE & Stack more than one AE after another & Perform slightly better than vanilla AE & Needs more resources than vanilla AE & Bearing vibration & \cite{tao2015bearing, roy2018stacked, galloway2016diagnosis} \\ 
SAE & AE constrained in training with sparsity to keeping neurons' activations low & Same as AE plus prevent overfitting by forcing all neurons to learn & More complex networks and need more resources than vanilla AE & Bearing vibration, turbine vibration & \cite{lu2017fault, chen2017rolling, galloway2016diagnosis, ahmed2018intelligent} \\ 
DAE & AE designed for noisy data & Outperform  vanilla AE with noisy data. Works slightly better stacking several DAEs & More complex networks and need more resources than vanilla AE; stacked DAE needs even more & Bearing vibration & \cite{lu2017fault, xia2017intelligent} \\  \hline
\multicolumn{6}{l}{\textbf{Generative}} \\* 
VAE & AE that maps input data to posterior distribution & Learns posterior distribution from noisy distribution, generating data non-deterministically & Difficulty on implementation. Loses temporal relations if input data are raw sensors data. & Ball screw, electrostatic coalescer, web traffic & \cite{wen2018degradationKDE, lygren2019unsupervised, xu2018unsupervised} \\ 
GAN & Used for data augmentation and AD in 2 ways: using discriminator and using residuals & Good data augmentation with small imbalance ratio. AD outperform unsupervised SotA methods & Not working well with big imbalance ratio, complex and need more resources. Outperformed by simpler methods as CNN \cite{cabezas2019generative} & Induction motor, bearing multisensor & \cite{cabezas2019generative, lee2017application} \\  \hline
\multicolumn{6}{l}{\textbf{One-Class Classifiers}} \\*
OC-NN & Train AE and freeze Encoder for OCC,  similar to OC-SVM loss function &  Automatic feature extraction & Slower than traditional OCCs. Extracted features are not focused on the problem & General AD & \cite{chalapathy2018anomaly} \\  \hline
\multicolumn{6}{l}{\textbf{Recurrent Neural Networks}} \\*
Vani-lla RNN & Regression, AD tracking error between predicted and real behavior or HI difference. & Model temporal relationships of time-series data. Self-learning. & Suffers vanishing gradient problem; therefore cannot model medium and long-term dependencies. Need more resources than feedforward AE or CNN for training. & Activity recognition & \cite{arifoglu2017activity} \\ 
LSTM & Same as vanilla RNN but changing neurons architecture to LSTMs &  Same as vanilla RNN, however these can model longer time dependencies than vanilla & Even if handle better the vanishing gradient problem than vanilla, have difficulty on modelling long-term dependencies. Long training and computational requirements & Aircraft data, activity recognition & \cite{nanduri2016anomaly, guo2016robust, arifoglu2017activity} \\ 
GRU & Same as vanilla RNN but changing neurons architecture to GRUs & Same as LSTM plus easier to train & Same as LSTM but obtain a little worse results & Aircraft data, activity recognition & \cite{nanduri2016anomaly, arifoglu2017activity} \\ \hline

\end{longtable}
}

\subsection{Diagnosis}

The diagnosis steps depends on the information and type of AD model used for the previous stage, given PdM is an incremental process where each stage is complemented by previous stages. In the case of multi-class classifier, the type of failure related to the detected anomaly is already known, which enables a straightforward diagnosis and comparison with historical data \cite{rad2011artificial, al2010rolling}. Nonetheless, most PdM architectures implement binary classifier, one-class classifier or unsupervised models, which lack of failure type information. Therefore, these can only perform diagnosis by grouping the detected anomalies among them by similarity, which is done using clustering models \cite{xu2018roller, aytekin2018clustering, zong2018deep, amarbayasgalan2018unsupervised} and SOM \cite{li2018data, rustum2017fault, schwartz2020fault, hao2017health}. The features used for this stage are similar to the ones for AD, which can be based either on traditional or deep learning techniques. 

\subsection{Prognosis}

The deep learning based models for PdM prognosis are focused on fitting a regression model to prognosticate either the remaining useful life (RUL) of the diagnosed failure or the health degradation when there is no historical data of that type. The RUL is commonly measured in time or number of cycles and the health degradation is tracked using anomaly deviation quantification by health indexes. The most common algorithms are summarised and compared in Table \ref{tab:DL_alg_for_prognosis}. Their input can be the information generated in previous stages and traditional or deep learning features. There are many other algorithms that use DL features or traditional features combined with fully-connected network as last layer to perform prognosis, but these are presented in the combination Section 3.6, while this section focuses on the most common and simple SotA techniques that only use DL for prognosis.

\begin{table}[!htbp]
\small 
\caption{Summary of DL based prognosis works for PdM. Unsup and sup in algorithm column refer to unsupervised and supervised respectively.}
\label{tab:DL_alg_for_prognosis}
\begin{tabular}{p{11mm}p{35mm}p{29mm}p{29mm}p{16mm}p{7mm}}
\hline
\textbf{Algo-rithm} & \textbf{How it works} & \textbf{Strengths} & \textbf{Limitations} & \textbf{Applica-tions} & \textbf{Ref} \\ \hline
Vanilla RNN Unsup and sup & Regression, predicting features' and HI's evolution or predicting remaining cycles or time. & Model temporal relationships of time-series data. Possibility of self-learning & Suffers vanishing gradient problem; therefore cannot model medium and long-term dependencies. High training and computational requirements & Aero engine & \cite{yuan2016fault} \\ 
LSTM Unsup and sup & Same as vanilla RNN but changing the neurons architecture to LSTMs & Same as vanilla RNN, however these can model longer time dependencies than vanilla. Outperform vanilla RNN & Even if handle better the vanishing gradient problem than vanilla, have difficulty on modelling long-term dependencies. High training and computational requirements & Aero engine, rolling bearing, lithium batteries & \cite{guo2017recurrent, yuan2016fault, zhang2019bearing, zhang2018long} \\ 
GRU Unsup and sup & Same as vanilla RNN but changing the neurons architecture to GRUs & Same as LSTM plus easier to train & Same as LSTM but obtain a little worse results & Aero engine, lithium batteries & \cite{yuan2016fault} \\ \hline

\end{tabular}
\end{table}

\subsection{Mitigation}

The research methodology followed to create the current publication, showed no DL-based mitigation publications. Several possible reasons for this fact are described bellow. The majority of DL works are focused on optimising a single performance metric for the ML task to be solved, like maximising accuracy or F1 score on classification, and minimising errors like MAE or RMSE on regressions. These works' solutions are usually compared in simulated reference datasets, looking for the architecture that outperforms the rest on the aforementioned metrics. Nonetheless, deep learning models are the hardest ML type to understand given their higher complexity that makes them more accurate at modelling high dimensionality complex data, and therefore they fail to meet the industrial explanation facility requirement.

In order to address this problem, they should provide mitigation advice or at least explanations about the reasons for making predictions, which could be supported on the emerging field explainable artificial intelligence (XAI). Furthermore, the final and most ambitious step in this PdM stage should be the automatising of recommendations for domain technicians to integrate PdM in the maintenance plan, by optimising industrial maintenance process via maintenance operation management. Finally, the reasons for existing few real application publications are presented bellow. Industrial companies avoid publishing their data or implementation details to protect their intellectual property and know-how from competence. Moreover, many data-driven research publications lack of domain technician feedback so they tackle the problem only relying on data-driven techniques, without embracing domain knowledge.

\subsection{Combination of models and remarkable works}

The DL techniques already presented throughout current section are the basic elements and architectures used for PdM. It is worth highlighting there are infinite possible architectures by combining these techniques among them, or used together with other data-driven or expert-knowledge based techniques. The combination and adaptation of models for the problem being addressed results into more accurate models that fulfil its requirements. Table \ref{tab:possible_combination_DL_alg_for_PdM} summarises how these models are commonly combined in SotA architectures, presenting their strengths and limitations. 

\begin{table}[!htbp]
\small 
\caption{Possible combination of deep learning techniques for PdM architectures.}
\label{tab:possible_combination_DL_alg_for_PdM}
\begin{tabular}{p{18mm}p{35mm}p{37mm}p{37mm}}

\hline

\textbf{Algorithm} & \textbf{How they work} & \textbf{Strengths} & \textbf{Limitations}
\\ \hline

\multicolumn{4}{l}{\textbf{Traditional and DL features combined with DL models}} \\ 
Traditional and DL-based FE with AE & Combine traditional and DL FE methods with already presented autoencoder architectures in the same model & Outperform traditional ML and simple DL architectures. No need of handcrafted features. Automatic FE. Can model time-series dependencies using CNN, LSTM and GRU by context extraction & Understanding deep features is not straightforward. Slower and more complex than simple ANN models \\ 
Traditional and DL-based FE with DBN & DL and traditional FE methods with DBN stacked to other models & Same as above & Same as above \\  \hline

\multicolumn{4}{l}{\textbf{Hybrid: combination of features and models}} \\ 
DL FE techniques combination & Combine CNN, LSTM, other DL FE techniques and traditional features to extract more complex features & Automatic dimension reduction. Outperform other FE techniques. Model temporal relations and neighbors. With bidirectional RNNs, future context is available & More complex and need more resources than traditional ML and simple DL models. Bidirectional RNN cannot be done online \\ \hline

\end{tabular}
\end{table}

Moreover, Table \ref{tab:combination_DL_alg_for_PdM_relevant_works_summary} contains relevant works of the aforementioned types, which merge traditional FE or deep learning FE with traditional data-driven or deep learning models. This collection of works shows that combination of techniques can address all PdM stages using supervised or unsupervised approaches.

{\small 
\setlength\LTleft{0mm} 
\begin{longtable}{p{18mm}p{28mm}p{29mm}p{29mm}p{23mm}}
\caption{Combination of deep learning techniques for PdM: relevant works summary.} \label{tab:combination_DL_alg_for_PdM_relevant_works_summary} \\ \hline
\textbf{Architecture} & \textbf{How it works} & \textbf{Strengths} & \textbf{Limitations} & \textbf{Applications and refs} \\ \hline
\endfirsthead
\nobreakhline
\textbf{Architecture} & \textbf{How it works} & \textbf{Strengths} & \textbf{Limitations} & \textbf{Applications and refs} \\* \nobreakhline
\endhead
\multicolumn{5}{l}{\textbf{Autoencoders}} \\*
AE with extreme learning machine (ELM). & Unsupervised AD tracking error of ELM for OCC, trained with normal data. & Two steps training. Easy to train. & Unable to model non-linear or complex relations in ELM. & Power plant \cite{michau2020ELMwithAE}, machine lifetime estimation \cite{bose2019adepos}. \\ 
Stacked SAE & Unsupervised FE adding noise & No need of preprocessing. Robust to noise. Severity identification. & Difficult optimisation of deep architecture & Rolling bearing \cite{chen2017rolling} \\ 
Stacked CNN-based AE  & Unsupervised FE modelling temporal relations in sliding window & Model temporality using neighbours. & Only short temporal relations & Gearbox vibration \cite{cabrera2017automatic} \\ 
AE with LSTM  & Unsupervised FE modelling temporal relations & Model temporality & Higher computational requirements & Aviation \cite{huang2019motor}, turbofan and milling machine \cite{malhotra2016multi}, solar energy, electrocardiogram \cite{pereira2018unsupervised} and manufacturing \cite{lindemann2019anomaly} \\ 
VAE with RNN, GRU or LSTM  & Unsupervised generative FE modelling temporal relations and reducing to latent gaussian distribution & Model temporality. Regularised latent space & High computational requirements & Motor vibration\cite{huang2019motor}, turbofan \cite{yoon2017semi}, sensors \cite{zhang2019time} \\ \nobreakhline
\multicolumn{5}{l}{\textbf{Restricted boltzmann machines and deep belief networks}} \\* 
DBN & Unsupervised FE by hierarchical representations & Fault classification from frequency distribution & Need preprocessing. Tendency to overfitting. Not modelling temporal relations & Induction motors fault simulator \cite{shao2017deep} \\  
Regularised RBM + SOM + RUL  & Probabilty modelling, health assesment and RUL prognosis using distance & RBM regularisation improve FE for RUL & Single RBM, can be improved by multiple of these layers. & Rotating systems \cite{liao2016enhanced} \\  
Image generation + DBN + MLP/FDA/SOM  & Supervised or unsupervised FE modelling from vibration image data & Model temporality in an image. Combine with image processing methods & Difficulty on extracting clusters' meaning, relying on domain knowledge. & Journal bearing \cite{oh2017scalable} \\ \nobreakhline
\multicolumn{5}{l}{\textbf{Hybrid: combination of features and models}} \\* 
Bidirectional LSTM  & Unsupervised FE modelling temporal relations & Health estimation and then RUL mapping. More robust. Future context is available. & Need all signal to be processed: no streaming. More complex than simple LSTM. & Turbofan \cite{elsheikh2019bidirectional} \\ 
AE + Convolutional DBN + exponential moving average (EMA) & Unsupervised probability modelling by automatic FE, modelling temporal relations. Training in steps & Model temporality. More stable than traditional ML and simple DL. Each model complement others weaknesses & Each part trained independently, not for problem. EMA only model shorter term temporal relations. & Electric locomotive bearing fault \cite{shao2017electric} \\ 
CNN and bidirectional LSTM based AE + fully connected + linear regression & Unsupervised FE modelling temporal relations & Raw sensor data modelling. Model long-term temporal dependencies & Sliding window needs complete window. Higher complexity combining DL techniques & Milling machine \cite{zhao2017learning} \\ 
Traditional FE + bidirectional GRU combined with ML models & Unsupervised FE modelling temporal relations & Same as above & Same as above & Aviation bearing fault detection, gear fault diagnosis and tool wear prediction \cite{zhao2017machine} \\ \hline

\end{longtable}
}

The rest of this subsection summarises the contributions and strengths of relevant analysed works. One interesting article was published by Shao et al. \cite{shao2017AEoptimisation}, where a methodology of AE optimisation for rotating machinery fault diagnosis is presented. Firstly, they create a new loss function based on maximum correntropy to enhance feature learning. Secondly, they optimise model's key parameters to adapt it to signal features. This model is applied to fault diagnosis of gearbox and roller bearing. Another relevant publication is by Lu et al. who use growing SOM \cite{lu2018data}, a extension of SOM algorithm that does not need specification of map dimension. It has been applied to simulated test cases with application in PdM.

Guo et al. \cite{guo2016robust} propose a model based on LSTM and EWMA control chart for change point detection that is suitable for online training. An additional interesting work is presented by Lejon et al. \cite{lejon2018machine}, who use ML techniques to detect anomalies in hot stamping machine by non-ML experts. They aim to detect anomalous strokes, where the machine is not working properly. They present the problem that most of the collected data corresponds to press strokes of products without defects and that all the data is unlabelled. This data comes from sensors that measure pressures, positions and temperature. The algorithms they benchmarked are AE, OCSVM and IF, where AE outperforms the rest achieving the least number of false positive instances. As the authors conclude, \textit{the obtained results show the potential of ML in this field in transient and non-stationary signals when fault characteristics are unknown}, adding that AEs fulfill the requirements of low implementation cost and close to real-time operation that will lead to more informed and effective decisions.

As previously mentioned in this article, the possibility of model combination is infinite. For instance, Li et al. in the work \cite{li2018anomaly} combine a GAN structure with LSTM neurons, two widely used DL techniques that achieve SotA results. Additionally, DL techniques can be combined with other computing techniques as Unal et al. do in \cite{unal2014fault}, combining a feed forward network with Genetic Algorithms.

The last highlighted article that combines DL models is by Zhang et al. \cite{zhang2019deep}, one of the most complete unsupervised PdM works. They build a model that uses correlation of sensor signals in the form of signature matrices as input that is fed into an AE that uses CNN and LSTM with attention for AD, partial RCA and RUL. The strengths of this work are the following: they show that correlation is a good descriptor for time-series signals, attention mechanism using LSTMs gives temporal context and the use of anomaly score as HI is useful for RCA, mapping the detected failures to the input sensors that originate them. Conversely, the RCA they do is not complete since they only correlate failures to input sensors but are not able to link them to physical meaning. Moreover, the lack of pooling layers together with the combination of DL techniques results in a complex model that is computationally expensive, needs more time and data for training and its decisions are hard to explain.

The following publications use other ML tasks combined with DL models for PdM, and other DL techniques. Wen et al. \cite{wen2017TransferLearningSAE} use transfer learning with a SAE for motor vibration AD, outperforming DBNs. The article by Wen et al. \cite{wen2019timeseriesADwithCNN} proposes a transfer learning based framework inspired in U-Net that is pretrained with univariate time-series synthetic data. The aim of this network is to be adaptable to other univariate or multivariate anomaly detection problems by fine-tuning.

Martinez et al. \cite{martinez2019ActiveLearningBayesianCNN} present a bayesian and CNN based DL classifier for AD. They first use a small labelled dataset to train the model. Then, the model is used to classify the remaining data and then, it uses uncertainty modelling to analyse the observations that cannot be correctly classified due to high entropy. Finally, it selects the top 100 with highest entropy to query an domain knowledge technician, asking him/her to label them in order to retrain the model with this new data. This procedure is followed until the model obtains a good accuracy. This work is an example of how to use two interesting techniques in the field of PdM to address the problem of lacking labelled data by querying domain technicians, showing them the instances from which the model can learn the most. Concretely, the aforementioned techniques belong to semi-supervised classification type using active learning. Similarly, the review by Khan et al. \cite{khan2018review} mentions that expert knowledge can help troubleshooting the model and, if domain technicians are available, the model could learn from them using a ML training technique called active learning where the model queries them in the learning stage. Moreover, Kateris et al. present the work \cite{kateris2014OC-SOMActiveLearning} where they use SOM as OCC model for AD together with active learning, to progressively learn different stages of faults.

Another interesting technique with PdM applications is deep reinforcement learning. The publication by Zhang et al. \cite{zhang2018HIDeepReinforcementLearning} uses it for HI learning, outperforming feed-forward networks but underperforming CNN and LSTM for AD and RUL. This technique consists of transferring the knowledge adquired from one dataset to another one. The procedure consists of reusing a part or the whole pretrained model adapting it to new's requirements, which sometimes requires retraining the model but this needs less data and time. 

\subsection{Related review works summary}

This subsection summarises the most relevant information of the review works related to this survey, highlighting their main contributions, detected challenges and gaps in the SotA works and their conclusions. 

The survey by Chalapathy and Chawla \cite{chalapathy2019deep} analyses the SotA DL approaches to address anomaly detection. The work by Rieger et al. \cite{rieger2019fast} makes a qualitative review on the SotA fast DL models applied for PdM in industrial internet of things (IIoT) environments. They argue that real-time processing is essential for IoT applications, meaning that a high latency system can lead to unintentional reactive maintenance due to insufficient time to plan maintenance. Moreover, they highlight how DL models can be optimised. They state that weight-sharing on RNNs enables parallel learning, which can help learning these type of nets that achieve SotA results in most PdM applications. Accordingly, they also justify the use of max-pooling layers when dealing with CNNs to eliminate redundant processing and thus optimise them. There are two DL reviews applied to fields that can be extrapolated to PdM: DL models for time series classification by Fawaz et al. \cite{fawaz2019deep} and DL to model sensor data \cite{wang2019deep}.

The review by Zhao et al. \cite{zhao2019deep} explains there are algorithms that use traditional and hand-crafted features whereas others use DL features for the problem, and presents the most common FE methods for DL based PdM systems. They state that both aforementioned features work properly in DL models, supported on their SotA revision. Many of these works use techniques to boost model performance as data-augmentation, model design and optimisation for the problem, adopting architectures that already work in the SotA. They also adapt the learning function and apply regularisations and tweak the number of neurons, connections, apply transfer learning or stack models in order to enhance model generalisation and prevent overfitting. The advantage of traditional and hand-crafted features is they are not problem specific, being applicable to other problems. Moreover, they are easy to understand by expert-knowledge technicians given that they are based on mathematical equations. However, as they are not problem specific, in some cases DL-based FE techniques perform better since these are learned specifically for the problem and directly from the data. However, they are not as intuitive as aforementioned features, meaning that technicians can have problems trying to understand how they work.

The article by Zhao et al. \cite{zhao2019deep} also summarises the information already stated throughout this survey: DL models can achieve SotA results, pre-training in AEs can boost their performance, denoising models are beneficial for PdM because of the nature of sensor data and that CNN and LSTM variants can achieve SotA results in the field of PdM using model-optimisation, depending on the dataset's scale. In addition, domain knowledge can help in FE and model optimisation. Conversely, it is difficult to understand DL models even if there are some visualisation techniques because they are black-box models. Transfer learning could be used when having little training data, and PdM belongs to a imbalaced class problem because faulty data is scarce or missing.

The survey by Zhang et al. \cite{Zhang2019} compares the accuracy obtained by ANN, Deep ANN and AE in different datasets, which allows comparisons, however these comparisons are done with models applied to different datasets and therefore they are not fair. Nonetheless, they show high accuracy results, most of them between 95\% and 100\%, emphasising that DL models can obtain promising results. They state that deeper models and higher dimensional feature vectors result in higher accuracy models but sufficient data is needed. With the increase of computational power and data growth in the field of PdM, research on this area tends to focus on data-driven techniques and specifically DL models. However, DL models lack of the explainability and interpretability of taken decisions.

The review by Khan et al. \cite{khan2018review} states that the developed DL architectures are application or equipment specific and therefore there is no clear way to select, design or implement those architectures; the researches do not tend to justify the decision of selecting one architecture over another that also works for the problem, for instance selecting CNN versus LSTM for RUL. Its authors also argue that SotA algorithms as the ones presented throughout this section all have shown to be working correctly and are not different from one another.

Even if this section has been focused on DL models for PdM, we have seen that they are often integrated with traditional models and/or traditionally FE features, such as time and frequency domains, feature extraction based on expert knowledge or mathematical equations.

As the authors Khan et al. state in \cite{khan2018review}, there is a lack of understanding of a problem when building DL models. They also argue that VAE is ideal for modelling complex systems, achieving high prediction accuracy without health status information. The algorithms that analyse the data maintaining its time-series relationship by analysing the variables together, at the same time, are the most successful: no matter if using sliding window, CNN or LSTM techniques. Most of SotA algorithms focus on AD, whereas they can also be adapted to perform RUL by a regression or RNN, where the majority use LSTMs. Regressions commonly use features learned for the used AD models, or even use traditional and hand-crafted features. Generative models like GAN do not work as good as expected. However, CNN works well while needing less data and computing effort. This means that even DL models can achieve similar accuracy using traditional features or deep features extracted from the data unsupervisedly.

\section{Comparison of state-of-the-art results} \label{sec:sota-benchmark-datasets}

\subsection{Benchmark datasets}

The review made by Khan et al. \cite{khan2018review} states that one of the problems of PdM proposals is the lack of benchmarking among them. There are some public PdM databases among the prognosis datasets released by the Nasa \cite{NASADatasetsProgno2018} belonging to the scope of predictive maintenance, which are presented in the following paragraphs.

\textit{3. Milling dataset} \cite{NASADatasetsProgno2018} gathers acoustic emission, vibration and current sensor data under different operating conditions with the purpose of analysing the wear of the milling insert. Regarding PdM stages, it allows the application of AD, RCA and RUL.

\textit{4. Bearing dataset} \cite{NASADatasetsProgno2018} gathers vibration data from 4 accelerometers that monitor bearings under constant pressure until failure, obtaining a run-to-failure dataset where all failures occur after exceeding their design life of 100 million revolutions. Its possible PdM applications are AD and RUL estimation.

\textit{6. Turbofan engine degradation simulation dataset} \cite{NASADatasetsProgno2018} contains run-to-failure data from engine sensors. Each instance starts at a random point of engine life where it works correctly, and monitors its evolution until an anomaly happens and afterwards reaches the failure state. The engines are working under different operational conditions and develop different failure modes. Its possible PdM applications are AD, RCA and RUL.

\textit{10. Femto bearing dataset} \cite{NASADatasetsProgno2018} is a bearing monitoring dataset inside the Pronostia competition that contains run-to-failure and sudden failure data. The used sensors are thermocouples gathering temperature data and accelerometers that monitor vibrations in the horizontal and vertical axis. Its possible PdM applications are AD, RCA and RUL.

Industrial companies are reluctant to publish their own datasets because they tend to trade secret their data and knowledge in order to protect themselves from their competence. The dataset that approximates most to companies data is the one published by Semeion research center named \textit{Steel plates faults dataset} \cite{UCI-ML-repo-datasets}, where steel plate faults are classified into 7 categories.

\subsection{Data-driven technique's results comparison}

This subsection compares different relevant data-driven works for PdM application turbofan dataset introduced in previous subsection, which is generated using the \textit{Commercial modular aero-propulsion system simulation (C-MAPPS)}. The reasons for choosing this dataset are that it is one of the reference datasets of PdM, it enables the application of all PdM steps and it is one of the most used dataset for model ranking.

The dataset lacks of the RUL label, which is the target column. Hence, many works assume it to be constant in the initial period of time where the system works in correct conditions and degrades linearly after exceeding the changepoint or initial anomalous point. The constant value in initial period is a parameter denominated as $R_{max}$, which is set to values near 130 for many state-of-the-art works, enabling a fair comparison of their results.

The most common metrics to evaluate models' performance are the following ones \cite{babu2016RULwithCNN}: root mean square error (RMSE) in Equation \ref{eq:RMSE}, and score function that penalises late predictions in Equation \ref{eq:scoring_function}, which was used in the PHM 2008 data challenge \cite{Saxena2008turbofanPHM}. In previous equations, $N$ is the number of engines in test set, S is the computed $score$, and $h=(Estimated RUL-True RUL)$.  Table \ref{tab:turbofan_results} gathers state-of-the-art results for the last years on the four subsets of the dataset.

\begin{equation} \label{eq:RMSE}
    \begin{aligned} RMSE = \sqrt{\frac{1}{N}\sum _{i=1}^{N}h_i^2} \end{aligned}
\end{equation}

\begin{equation} \label{eq:scoring_function}
    \begin{aligned} S = \left\{ \begin{array}{cc} \sum \nolimits _{i=1}^{N} \left( e^{-\frac{h_i}{13}}-1\right) &{} for~~h_i < 0 \\ \sum \nolimits _{i=1}^{N} \left( e^{\frac{h_i}{10}}-1\right) &{} for~~h_i \ge 0 \end{array}\right. \end{aligned}
\end{equation}

\begin{table}[!htbp]
\small 
\caption{State-of-the-art results on four turbofan dataset subsets since 2014. The lower the metric, the better the model is considered to perform on average. Best results are highlighted in bold.}
\label{tab:turbofan_results}
\begin{tabular}{p{22mm}p{5mm}p{5mm}p{18mm}p{7mm}p{7mm}p{7mm}p{7mm}p{8mm}p{8mm}p{8mm}p{8mm}}
\hline

\textbf{Reference} & \textbf{Year} & \textbf{$R_{max}$} & \textbf{Architecture} & \textbf{FD001 RMSE} & \textbf{FD002 RMSE} & \textbf{FD003 RMSE} & \textbf{FD004 RMSE} & \textbf{FD001 Score} & \textbf{FD002 Score} & \textbf{FD003 Score} & \textbf{FD004 Score}  \\ \hline
Ramasso et al. \cite{ramasso2014performance} & 2014 & 135 & RULCLIPPER & 13.3 & 22.9 & 16.0 & 24.3 & \textbf{216} & 2796 & 317 & 3132  \\ 
\multirow{4}{*}{Babu et al. \cite{babu2016RULwithCNN}} & \multirow{4}{*}{2016} & \multirow{4}{*}{130} & MLP & 37.6 & 80.0 & 37.4 & 77.4 & 17972 & 7802800 & 17409 & 5616600  \\ 
 &  & & SVR & 21.0 & 42.0 & 21.0 & 45.3 & 1381 & 589900 & 1598 & 371140  \\ 
 &  & & RVR & 23.8 & 31.3 & 22.4 & 34.3 & 1504 & 17423 & 1431 & 26509  \\ 
 &  & & DCNN & 18.4 & 30.3 & 19.8 & 29.2 & 1287 & 13570 & 1596 & 7886  \\ 
Zhang et al. \cite{zhang2016multiobjective} & 2017 & 130 & MODBNE & 15.0 & 25.1 & 12.5 & 28.7 & 334 & 5585 & 422 & 6558  \\ 
Zheng et al. \cite{zheng2017long} & 2017 & 130 & LSTM + FFNN & 16.1 & 24.5 & 16.2 & 28.2 & 338 & 4450 & 852 & 5550 \\ 
Li et al. \cite{li2018remaining} & 2018 & 125 & CNN + FFNN & \textbf{12.6} & 22.4 & 12.6 & 23.3 & 273 & 10412 & 284 & 12466  \\ 
Ellefsen et al. \cite{ellefsen2019remaining} & 2019 & 115-135 & RBM + LSTM & \textbf{12.6} & 22.7 & \textbf{12.1} & 22.7 & 231 & 3366 & 251 & \textbf{2840}  \\ 
Da Costa et al. \cite{da2019attention} & 2019 & 125 & LSTM+attention & 14.0 & \textbf{17.7} & 12.7 & \textbf{20.2} & 320 & \textbf{2102} & \textbf{223} & 3100  \\ \hline

\end{tabular}
\end{table}

Results comparison of Table \ref{tab:turbofan_results} does not show only model's performance, but also the combination of preprocessing and feature engineering techniques. Therefore, results show the performance of the whole data process applied to the dataset until prediction. Nonetheless, the table shows that deep learning based architectures are the ones that achieve state-of-the-art results in recent years. Concretely, these architectures are composed of combination of different DL techniques.

\section{Discussion}

This section analyses deep learning architectures' suitability in the field of PdM. It is the result of comparing reviewed articles' trends, results and conclusions with PdM data characteristics and industrial requirements.

\subsection{Comparison and suitability of deep learning in predictive maintenance}

Physical and knowledge-based models for PdM were widely used 15 years ago but they are less common nowadays due to the difficulty or impossibility of modelling complex systems. In fact, data-driven statistical and machine learning publications started to gain popularity in this field since they learn system's behaviour from the data directly and therefore needed little domain knowledge. Conversely, in later years, due to the emergence of I4.0, the increment of computational power and the automatising of machine and asset data collection, the data-driven publication trend has moved towards deep learning based schemes.

There are several reasons for deep learning being a hot research topic in predictive maintenance field. They usually achieve higher accuracy than traditional data-driven techniques. They can dispense with expert knowledge feature engineering given their capacity of extracting automatic features for the problem being addressed. In addition, they can model time-series data using attention or time context. The application of DL models is also widely researched in other fields such as image processing and seq2seq. Nonetheless, their two major drawbacks are high training data requirements and difficulty on model explainability. Conversely, these models must be modified and adapted for industrial and PdM data characteristics and requirements. 

Therefore, the model type choice for PdM application should be done carefully, after analysing each use case's needs. Maybe, its requirements are not satisfied by that moment's machine learning research trend, which is currently deep learning, and other type of models are more appropriate. For instance, statistical, machine learning and deep learning models have their own peculiarities. They are all able to fulfill the following PdM desirable characteristics from the list \cite{venkatasubramanian2003review} by creating specific architectures: quick detection and diagnosis, isolability, novelty identifiability, classification error estimation, adaptability, and real-time computation and multiple fault identifiability. However, the main differences among these type of models are summarised in Table \ref{tab:DL-ML-statistics_forPdM_comparison}. Hence, the election of one group over the rest and even deciding the final architecture requires a thorough analysis and comparison to determine the one that suits both: use case and its data requirements.

\begin{table}[!htbp]
\small 
\caption{Differences of statistical, machine learning and deep learning architectures for predictive maintenance.}
\label{tab:DL-ML-statistics_forPdM_comparison}
\begin{tabular}{p{35mm}p{30mm}p{30mm}p{30mm}}

\hline

\textbf{Characteristic} & \textbf{Statistical} & \textbf{Machine learning} & \textbf{Deep learning}  \\ \hline
Amount of data for training & Small & Medium & High \\ 
Training time & Small & Medium & High \\ 
Complexity & Small & Medium & High \\ 
\multirow{2}{*}{Explanation facility} & \multirow{2}{*}{High} & Medium (grey models) & \multirow{2}{*}{Low} \\
& & Low (blackbox models) & \\ 
Accuracy & Low & Medium & High \\ \hline

\end{tabular}

\end{table}

In the end, most deep learning architectures are either based on traditional data-driven concepts or are combined with them in order to fill their gaps. Therefore, DL models could be a piece inside a PdM architecture that combines other kind of models presented in Table \ref{tab:DL-ML-statistics_forPdM_comparison}. This could compensate the drawback of some models with others by a fusion that meets PdM needs better.

\subsection{Automatic development of deep learning models for predictive maintenance}

Even though deep learning models can achieve SotA results in PdM datasets, their design, development and optimisation relies on publications, data scientists previous knowledge and trial and error testing. These are some of their biggest challenges: architecture type and structure choice, number of hidden layers and neurons, activation functions, regularisation terms to prevent overfitting and learning parameters optimisation.

For the above-stated reasons, the whole process of DL model creation is not as automatic as believed. Moreover, in order to obtain competitive results, many authors preprocess and feature engineer the raw EOC signals. This can boost model performance but at the same time remove relevant information that could be learnt automatically using more complex architectures. In addition, these steps are commonly performed by data scientists. Usually, domain knowledge is not embedded, so models are expected to learn all the non-linear relations from the data. Conversely, this information could help in architecture's dimensionality reduction, resulting in simpler, more accurate and as a result explainable models. Other byproduct benefits are less training data requirements, less training time and higher generalisation to avoiding overfitting.

\subsection{Application of deep learning research in industrial processes}

There are many works that apply deep learning for predictive maintenance in the literature. Most SotA deep learning techniques tackle PdM unsupervisedly given the difficulty to obtain failure data in industrial companies. This is the reason for AEs, RBMs and generative models having so much repercussion in the field. The following paragraphs summarise common techniques and how they meet industrial requirements. 

Regarding SotA, there are many DL proposals for AD and RUL. Most of them tend to combine different algorithms to create a more complex model that contains advantages of the techniques that compromise it. The most common combination for PdM sensor modelling in unsupervised way is CNNs with LSTMs in an AE or derived architecture. Similarly, supervised approaches usually use CNNs and LSTMs in a ANN that outputs probability of failure types or regressions. However, techniques fusion augments model complexity. 

Regarding the diagnosis step, it is easy to perform RCA with supervised models given that, when the training data contains the label, failure or not or even the type of failure, the model can directly map the new data with the corresponding failure type automatically. However, in companies that lack this type of data, they can only model normality by OCC models or even use unsupervised approach to model unlabelled data. There is a gap in these latter models since they are unable to perform complete RCA given the impossibility to classify unspecified failure types. One underlying reason could be the lack of collaboration between data scientists and expert-knowledge technicians. Therefore, this gap could be filled by applying explainable artificial intelligence (XAI) techniques to facilitate the communication, understanding and guidance of DL models. XAI is a promising emergent field with few publications in the field of PdM.

Deep learning models also fail to propose mitigation actions since, as mentioned before, they should work together with domain technicians knowledge and many works do not, tackling the problem in a purely data-scientific way and forgetting about the underlying process working knowledge. For this reason, even if many models are accurate, they can not meet industrial and real PdM requirements. They present complex schemes with many hidden layers even if Venkatasubramanian et al. \cite{venkatasubramanian2003review} state that understandability is one desired characteristic for PdM models. Without it, industrial companies may not deploy a deep learning models to production as domain technicians would be unable to understand their predictions and therefore trust them. Once again, the application of XAI techniques together with expert knowledge could overcome the problem by enabling to: understand the predictions, map detected failures to real physical root cause and even propose mitigation actions giving data-driven advice to help in maintenance management (MM) and manufacturing operation management (MOM) decision making.

The majority of reviewed works were created and tested in research environments but not transferred or tested in industrial companies. Even if there are some models trained with real industrial process data, the majority use reference datasets that have been preprocessed and specifically prepared for the task, such as the ones presented in Section \ref{sec:sota-benchmark-datasets} that are generated in simulation or testing environments. However these are unable to adapt to industrial companies' requirements presented by Venkatasubramanian et al. in article \cite{venkatasubramanian2003review} that still prevail nowadays. Lejon et al. in the work \cite{lejon2018machine} consolidate the aforementioned needs by stating that industrial data is unlabelled and mostly correspond to non-anomalous process conditions. With regard to PdM architectures, the work by Khan et al. \cite{khan2018review} seems to be the one that summarises and could better fit the requirements of the companies, even though it lacks of specification on how to address PdM in real companies.

All in all, we have seen that industrial companies need PdM models to be accurate, easy to understand, process data on streaming and adapted to process data characteristics. Their data is mostly collected in unsupervised way, or only non-failure data is available. Moreover, it is collected under different EOC. Conversely, there is a gap in the published data-driven models because available unsupervised and OCC proposals are unable to link novel detected failures to their physical meaning. The main reason is that these models ignore expert knowledge. In addition, there are few research publications on the application of XAI techniques in PdM, which could provide solutions for the main presented gaps.

\section{Conclusions}

The majority of industrial companies that rely on corrective and periodical maintenance strategies can optimise costs by integrating automatic data-driven predictive maintenance models. These models monitor machine and component states, whose research has evolved from statistical to more complex machine learning techniques. Nowadays, their main research focuses on deep learning models.

The main objective of this survey is to analyse the state-of-the-art deep learning techniques implementation in the field of predictive maintenance. For that purpose, several analysis and research are reviewed throughout the work, which are summarised in this paragraph. In the beginning, the most relevant factors and characteristics of industrial and PdM datasets are presented. Secondly, the steps necessary to perform PdM are presented in a methodological way. Afterwards, statistical and traditional machine learning techniques for PdM are reviewed, in order to gain knowledge on baseline models in which some deep learning implementations are based. Thenceforth, a thorough review on deep learning state-of-the-art works is performed, classifying the works by their underlying technique, data typology and compared among them; which enables methods' comparison in a structured way. Related reviews on DL for PdM are also analysed, highlighting their main conclusions. Thereafter, a summary on the main public PdM datasets is presented and SotA results are compared on turbofan engine degradation simulation dataset. Moreover, the suitability and impact of deep learning in the field of predictive maintenance is presented, together with the comparison with other data-driven methods. In addition, the systematisation of deep learning models development for predictive maintenance is discussed. Finally, the application of these models in real industrial use-cases is argued, analysing their applicability beyond public benchmark datasets and research environments. 

As stated before, industrial companies that want to optimise their maintenance operations should transition towards predictive maintenance. However, this automatising should be embraced from simpler to more complex models, always choosing the ones that could better fit their specific needs. Both domain experts and data scientists should collaborate in the development and validation of a PdM structure. This hybrid model could benefit from the advantages of both domain knowledge-based and data-driven approaches, resulting in an accurate yet interpretable model. Explainable machine learning applied to deep learning could be an alternative to white-box and grey box models, which are more interpretable and less accurate. These new models may achieve a trade-off between accuracy and explainability, integrating with domain knowledge technicians, which can use them as a tool to perform PdM and gain knowledge from the data while contrasting with theoretical background and domain expertise.

Industrial companies nowadays have collected much data by monitoring assets under normal working condition and little to none failure data. Therefore, unsupervised and one-class classification algorithms research is relevant for predictive maintenance field. Concretely, architectures like autoencoders or deep belief networks with LSTMs or CNNs are one of the most researched type of architecture that enable unsupervised time-series data modelling. Nonetheless, the design and optimisation of DL architectures is mainly guided by previous experience and trial and error. 

To sum up, deep learning models have gained popularity in PdM due to their high accuracy, achieving state-of-the-art results when trained with enough data. However, many works do not address other relevant aspects for PdM models such as interpretability, real time execution, novelty detection or uncertainty modelling, given that mainly laboratory datasets have been used. These aspects are fundamental to transfer any machine learning model to real, industrial use cases, and run in production.

\bibliographystyle{ACM-Reference-Format}
\bibliography{bibliography}

\end{document}